\documentclass[preprint, authoryear,12pt]{elsarticle}

\usepackage{amssymb}

\graphicspath{{}}
\DeclareGraphicsExtensions{.pdf,.jpeg,.png,.eps}

\usepackage[cmex10]{amsmath}
\usepackage[caption=false,font=normalsize,labelfont=sf,textfont=sf]{subfig}

\usepackage{fixltx2e}
\makeatletter
\def\@outputdblcol{%
  \if@firstcolumn
    \global\@firstcolumnfalse
    \global\setbox\@leftcolumn\copy\@outputbox
    \splitmaxdepth\maxdimen
    \vbadness\maxdimen
    \setbox\@outputbox\vbox{\unvbox\@outputbox\unskip}%
    \setbox\@outputbox\vsplit\@outputbox to\maxdimen
    \toks@\expandafter{\topmark}%
    \xdef\@firstcoltopmark{\the\toks@}%
    \toks@\expandafter{\splitfirstmark}%
    \xdef\@firstcolfirstmark{\the\toks@}%
    \ifx\@firstcolfirstmark\@empty
      \global\let\@setmarks\relax
    \else
      \gdef\@setmarks{%
        \let\firstmark\@firstcolfirstmark
        \let\topmark\@firstcoltopmark}%
    \fi
  \else
    \global\@firstcolumntrue
    \setbox\@outputbox\vbox{%
     \hb@xt@\textwidth{%
        \hb@xt@\columnwidth{\box\@leftcolumn \hss}%
        \hfil
        \vrule \@width\columnseprule
        \hfil
       \hb@xt@\columnwidth{\box\@outputbox \hss}}}%
  \@combinedblfloats
    \@setmarks
    \@outputpage
    \begingroup
      \@dblfloatplacement
      \@startdblcolumn
      \@whilesw\if@fcolmade \fi{\@outputpage\@startdblcolumn}%
    \endgroup
  \fi}
\makeatother

\makeatletter
\def\ps@pprintTitle{%
  \let\@oddhead\@empty
  \let\@evenhead\@empty
  \def\@oddfoot{\reset@font\hfil\thepage\hfil}
  \let\@evenfoot\@oddfoot
}
\makeatother

\usepackage{stmaryrd}
\usepackage{multirow}
\usepackage{nohyperref}
\journal{Pattern Recognition Letters}

\begin{document}

\begin{frontmatter}



\title{Shedding Light on the Asymmetric Learning Capability of AdaBoost}


\author[uvigo]{Iago Landesa-V\'azquez}
\ead{iagolv@gts.uvigo.es}

\author[uvigo] {Jos\'e Luis Alba-Castro}
\ead{jalba@gts.uvigo.es}

\address[uvigo]{Signal Theory and Communications Department, University of Vigo, Maxwell Street, 36310, Vigo, Spain}

\begin{abstract}
In this paper, we propose a different insight to analyze AdaBoost. This analysis reveals that, beyond some preconceptions, AdaBoost can be directly used as an asymmetric learning algorithm, preserving all its theoretical properties. A novel class-conditional description of AdaBoost, which models the actual asymmetric behavior of the algorithm, is presented.
\end{abstract}


\end{frontmatter}



\section{Introduction}

Asymmetry is present in many real world pattern recognition applications. Medical diagnosis, disaster prediction, biometrics, fraud detection, etc. have obviously different costs associated with the different kinds of mistakes (false positives and false negatives) implicitly related to each decision. But asymmetry is not only connected to the direct cost of a mistake. Many problems have unbalanced class priors, where one of the classes is extremely more frequent than the other one, or it is easier to sample. This kind of problems may require classifiers capable of focusing their attention in the rare (but most valuable) class, instead of trying to find hypothesis that in general fit well to data (mainly driven by the prevalent class).

From its original publication, boosting algorithms \citep{Schapire90} and specifically AdaBoost \citep{FreundSchapire97} have drawn a lot of attention of the pattern recognition community. Its strong properties and theoretical guarantees, its tendency to non-overfitting and its promising practical results, have focused the interest in this family of algorithms (e.g., \citealp{Schapire98, SchapireSinger99, Friedman00, MeaseWyner08a, ViolaJones04}) both from the theoretical (different interpretations, modifications, discussions\ldots) and practical points of view.

In the literature, several modifications of AdaBoost have been proposed to deal with asymmetry \citep{KarakoulasShawe98, Fan99, Ting00, ViolaJones04, ViolaJones02, Sun07,  MasnadiVasconcelos07}. Viola and Jones in their face detector framework (2004), use a validation set to modify the AdaBoost strong classifier threshold in order to trade off false positive and detection rates. Nevertheless, as they stated, it is not clear whether this change preserves the original training and generalization guarantees of AdaBoost \citep{ViolaJones04} and the weak classifiers selection is not optimal for an asymmetric task \citep{ViolaJones02}. Most of the other proposed algorithms \citep{KarakoulasShawe98, Fan99, Ting00, ViolaJones02, Sun07} try to reach asymmetry based on direct manipulations of the weight distribution update rule. These are heuristic modifications of the algorithm, but not a full reformulation of AdaBoost for asymmetric classification problems. On the other hand, the more recent Asymmetric Boosting algorithm \citep{MasnadiVasconcelos07} finds a new solution to the problem based on the Statistical View of Boosting \citep{Friedman00}. Their result is theoretically solid, but the final algorithm is far more complex and computing demanding than the original AdaBoost.

Eventhough some studies \citep{FreundSchapire97, Zadrozny03} mention that the incorporation of unbalanced initial weights could lead to a cost-sensitive version of AdaBoost, subsequent works insist that this is not enough to reach effective asymmetry \citep{ViolaJones02, Mease07, Sun07, MasnadiVasconcelos07} swelling the number of different asymmetric boosting algorithm variants. Meanwhile, standard AdaBoost remains being explained with an uniform initial weight distribution (e.g., \citealp{SchapireSinger99, Friedman00, Schapire98, Fan99, Ting00, Sun07, MasnadiVasconcelos07, FreundSchapire99, Polikar06, Polikar07}). To the best of our knowledge, a formal explanation of the consequences of using asymmetric initial weights on AdaBoost has not been provided, either in one way (they lead to effective asymmetry) or the other (they are definetely useless), so we think that some light must be shed in order to clarify the actual asymmetric learning capabilities of AdaBoost.

In this paper we propose a new perspective to analyze AdaBoost in a class-conditional way. This analysis suggests that, only with an unbalanced class-conditional initialization of the weight distribution, AdaBoost is, by itself, a theoretically sound asymmetric classification algorithm. Based on class error decomposition, our analysis offers a new model to understand AdaBoost behavior and how it really deals with asymmetry in an additive round-by-round scheme. In fact, weights initialization is no more than a way to modify the data distribution seen by the learner and, as we will see, it can be easily shown to shape the error bound that sets AdaBoost minimization goal. One key point of our work is that it is merely an analysis, so AdaBoost is unchanged. As a consequence, all the algorithm theoretical properties (related to training and generalization errors) remain intact, which has not been clearly reported on the other modifications in the literature. Our analysis is inspired by the generalized derivation of \citet*{SchapireSinger99}, close to the original \citep{FreundSchapire97} and specially intuitive and illustrative for our purpose. The Statistical View of Boosting \citep{Friedman00} and all its subsequent controversy \citep{MeaseWyner08a, Bennet08, MeaseWyner08b} is left aside, although an analogous conclusion could be derived from it.

The paper is organized as follows: in the next section we will describe AdaBoost original algorithm and its relationship with asymmetry. Section~\ref{sec:revisit} will detail our novel class-dependant interpretation, its analysis and some experimental results which show the actual asymmetric behavior of AdaBoost. Finally, Section~\ref{sec:conclusion} includes the main conclusions drawn from this analysis.

\section{AdaBoost}
\label{sec:AdaBoost}

In this section we will analyze the original AdaBoost definition and how it has usually been adapted to asymmetric learning.

\subsection{Algorithm}
\label{sub_sec:Algorithm_orig}

Given a set of $n$ training examples $(x_{i},y_{i})$ from which the $m$ first are positives $\{y_{i}=1\}_{i=1}^{m}$ and the rest are negatives $\{y_{i}=-1\}_{i=m+1}^{n}$, AdaBoost is a boosting algorithm whose goal is learning a strong classifier $H(x)$ based on an ensemble of weak classifiers $h_{t}(x)$ combined in a weighted voting scheme.

\begin{equation}
\label{strong_clas_eqn}
H(x)=
\mathrm{sign}\left(f(x)\right)=
\mathrm{sign}\left(\sum_{t=1}^T\alpha_{t}h_{t}(x)\right)
\end{equation}

To achieve this, a weight distribution $D_{t}(i)$ is defined over the whole training set. In each learning round $t$ the weak learner selects the best classifier according to the weight distribution, and this weak classifier is added to the ensemble weighted by a goodness parameter $\alpha_{t}$ (\ref{alphat_eqn}) depending on a correlation term $r_{t}$ (\ref{rt_eqn}). Once every weak classifier is selected, the weight distribution is updated according to its performance, following the rule on (\ref{weight_rule_eqn}) (where $Z_{t}$ is a normalization factor which ensures $D_{t}(i)$ is an actual distribution). The process can be repeated iteratively until a fixed number of rounds is reached, or when the obtained strong classifier achieves some performance goal.

\begin{equation}
\label{alphat_eqn}
\alpha_{t}=
\frac{1}{2} \ln \left(\frac{1+r_{t}}{1-r_{t}}\right)
\end{equation}

\begin{equation}
\label{rt_eqn}
r_{t}=
\sum_{i=1}^{n}D_{t}(i) y_{i} h_{t} (x_{i})
\end{equation}

\begin{equation}
\label{weight_rule_eqn}
D_{t+1}(i)=
\frac{D_{t}(i)\exp\left(-\alpha_{t} y_{i} h_{t}(x_{i})\right)}{Z_{t}}
\end{equation}

\begin{equation}
\label{zt_eqn}
Z_{t}=
\sum_{i=1}^{n} D_{t}(i) \exp\left(-\alpha_{t} y_{i} h_{t}(x_{i})\right)
\end{equation}


This framework can be seen \citep{SchapireSinger99} as an additive (round-by-round) minimization process of an exponential bound on the training error of the strong classifier. The bounding process is based on (\ref{bound_ineq_eqn}), and all the above expressions (including the weight update rule) can be derived from it.

\begin{equation}
\label{bound_ineq_eqn}
H(x_{i})\neq y_{i} \:\Rightarrow\: y_{i} f(x_{i}) \leq 0 \:\Rightarrow\: \exp\left(-y_{i} f(x_{i})\right) \geq 1
\end{equation}

Following the procedure used by \citet{SchapireSinger99}, the final bound of the training error obtained by AdaBoost is expressed as (\ref{et_bound_eqn}). The additive minimization of $\tilde{E}_{T}$ can be seen as finding, round by round, the weak hypothesis $h_{t}$ that maximizes $r_{t}$, that is maximizing the correlation between labels $(y_{i})$ and predictions $(h_{t})$ weighted by $D_{t}(i)$. 

\begin{equation}
\label{et_bound_eqn}
E_{T}\leq
\prod_{t=1}^{T}Z_{t}\leq
\prod_{t=1}^{T}\sqrt{1-{r_{t}}^2}=\tilde{E}_{T}
\end{equation}

For the sake of simplicity and clarity in our analysis, we will focus on the discrete version of the algorithm. In that case weak hypothesis are binary $y_{i}\in\{-1,+1\}$, and the minimization process is equivalent to selecting the weak classifier with less weighted error $\epsilon_{t}$ (\ref{round_err_eqn})\footnote{\emph{Notation}: Operator $\llbracket a \rrbracket$ is $1$ if $a$ is true and $0$ otherwise. The term `ok' refers to those training examples in which the result of the weak classifier is right $\{i:h(x_{i})=y_{i}\}$ and `nok' when it is wrong $\{i:h(x_{i})\neq y_{i}\}$.}. In this case, the last inequality on (\ref{et_bound_eqn}) becomes an equality, and parameter $\alpha_{t}$ can be rewritten (\ref{alphat2_eqn}) in terms of $\epsilon_{t}$ .

\begin{equation}
\label{round_err_eqn}
\epsilon_{t}=
\sum_{i=1}^{n} D_{t}(i) \llbracket h(x_{i}) \neq y_{i}\rrbracket=
\sum_{\textrm{nok}}D_{t}(i)
\end{equation}

\begin{equation}
\label{alphat2_eqn}
\alpha_{t}=
\frac{1}{2} \ln \left( \frac{1-\epsilon_{t}}{\epsilon_{t}}\right)
\end{equation}

This simplification doesn't prevent our analysis from being extended to other AdaBoost variations.

\subsection{AdaBoost and Asymmetry}
\label{sub_sec:AdaBoost_asymmetry}

\begin{figure}
\centering
\includegraphics[width=3.5in]{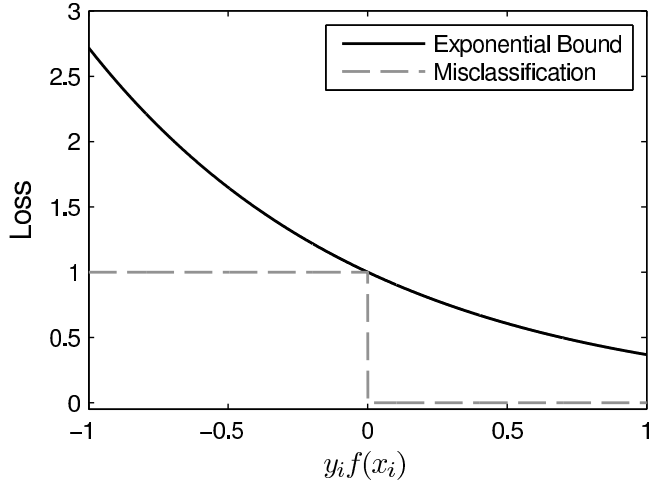}
\caption{AdaBoost exponential training error bound. Horizontal axis represents the absolute value of the final score of the strong classifier, with negative sign for errors and positive for correct classifications. Vertical axis is the loss related to misclassification and its exponential bound.}
\label{general_bound_fig}
\end{figure}

AdaBoost is usually seen as a learning procedure driven by misclassification on the training set. In that sense, the exponential bound to minimize must be defined (\ref{exp_bound_eqn}) following the guidelines proposed by \citep{SchapireSinger99}. Graphically, we can visualize this bounding process in Figure \ref{general_bound_fig}.

\begin{equation}
\label{exp_bound_eqn}
\begin{split}
E_{T}&=
\frac{1}{n}\sum_{i=1}^{n} \llbracket H(x_{i}) \neq y_{i}\rrbracket \\
&\leq \frac{1}{n}\sum_{i=1}^{n} \exp \left( -y_{i} f(x_{i}) \right)= 
\prod_{t=1}^{T} Z_{t} =
\tilde{E}_T
\end{split}
\end{equation}

From this point of view, AdaBoost is an algorithm with a symmetric behavior if the number of instances in the training set is the same for the two classes, or biased to the prevalent class otherwise. Consequently, AdaBoost couldn't be a cost-sensitive algorithm unless the training data set is resampled accordingly.

Based on this seemingly balanced nature, several modifications of AdaBoost have been proposed in order to adapt the algorithm to cost-sensitive problems. Most of them \citep{KarakoulasShawe98, Fan99, Ting00, Sun07, ViolaJones02} are based on modifying the weight update rule in an asymmetric (class-conditional) way. However it is not clear how these changes can affect the theoretical properties of AdaBoost since, as was mentioned above, the update rule is a consequence of the minimization process and not an arbitrary starting point of it.

This perspective is supported by the fact that AdaBoost is usually explained with a fixed uniform initial weight distribution ($D_{1}(i)=1/n$) (e.g., \citealp{SchapireSinger99, Friedman00, Schapire98, Fan99, Ting00, Sun07, MasnadiVasconcelos07, FreundSchapire99, Polikar06, Polikar07}). Nevertheless some initial works by \citet{FreundSchapire97} leave this distribution free to be controlled by the learner. In our explanation of the algorithm in  Section~\ref{sub_sec:Algorithm_orig} we deliberately didn't mention anything about the initialization of the weight distribution. So, what would really happen if the initial distribution was a generic one? Changes of the initial distribution were used to deal with cost-related utility functions \citep{Schapire98b}, and cost-sensitive weight initializations bonded to different changes in the weight update rule were also used by \citet{KarakoulasShawe98}, \citet{Fan99} or \citet{Ting00}. \citet{ViolaJones02} proposed a first modification of AdaBoost equivalent to an asymmetric modification of the initial weights. Nevertheless, they discard this approximation arguing that the induced asymmetry is fully absorbed by the first round, remaining the rest of the process entirely symmetric. Their final proposal (coined as \emph{AsymBoost}) was fairly spreading the desired asymmetry among a predefined number of rounds.

Though it is not widely appreciated, it can be easily shown that the error bounded and minimized by AdaBoost is actually a weighted error depending on the initial weight distribution. The only change with regard to the usual bound (\ref{exp_bound_eqn}), in which initial uniform weights have been taken out of the summation, is that generic initial weights must be kept inside the summation during the bounding process (\ref{exp_bound2_eqn}). 

\begin{equation}
\label{exp_bound2_eqn}
\begin{split}
E_{T}&=
\sum_{i=1}^{n} D_{1}(i) \llbracket H(x_{i}) \neq y_{i}\rrbracket \\
&\leq \sum_{i=1}^{n} D_{1}(i) \exp \left( -y_{i} f(x_{i}) \right)= 
\prod_{t=1}^{T} Z_{t} =
\tilde{E}_T
\end{split}
\end{equation}

All the rest of the process remains identical to that explained by \citet{SchapireSinger99}, consequently guaranteeing all the theoretical properties of AdaBoost with regard to training and generalization errors.

\section{Revisiting AdaBoost}
\label{sec:revisit}

In this section we will show our novel class-conditional interpretation model for AdaBoost. This generalized analysis will shed light on the class-dependant behavior of AdaBoost sketched in the previous section.

\subsection{Asymmetric Interpretation}
\label{sub_sec:asym_interp}

To derive our new interpretation of AdaBoost, instead of the initial weight distribution used in the original AdaBoost formulation, we define a set of parameters which contain exactly the same information of the former distribution. 

\vspace{3pt}
\begin{itemize}
\item \emph{Asymmetry}:
\begin{equation}
\label{asym_eqn}
\sum_{i=1}^{m} D_{1}(i) = \gamma\in (0,1)
\end{equation}
\item \emph{Class-conditional distributions}:
\begin{gather}
\label{pos_weight_ini_eqn}
D_{1+}(i)=
\frac{D_{1}(i)}{\gamma}, \quad \textrm{for } i=1,\ldots,m\\
\label{neg_weight_ini_eqn}
D_{1-}(i)=
\frac{D_{1}(i)}{1-\gamma}, \quad \textrm{for } i=m+1,\ldots,n
\end{gather}
\end{itemize}
\vspace{3pt}

If we put this new set of parameters into the training error expression (\ref{exp_bound2_eqn}) we will be able to decompose it in terms of its positive and negative class error components:

\begin{equation}
\label{et_classcond_eqn}
\begin{split}
E_{T}&=
\sum_{i=1}^{n} D_{1}(i) \llbracket H(x_{i}) \neq y_{i} \rrbracket =
\gamma \sum_{i=1}^{m} D_{1+}(i) \llbracket H(x_{i}) \neq y_{i} \rrbracket \\
&\quad + \left(1-\gamma\right) \sum_{i=m+1}^{n} D_{1-}(i) \llbracket H(x_{i}) \neq y_{i} \rrbracket \\
&= \gamma\, E_{T+} + \left(1-\gamma\right) E_{T-}
\end{split}
\end{equation}

Bounding (\ref{et_classcond_eqn}) with the usual exponential approximation, we can also obtain the error bound as the combination of two class-conditional partial error bounds:

\begin{equation}
\label{exp_bound_classcond_eqn}
\begin{split}
E_{T}&=
\gamma\, E_{T+} + \left(1-\gamma\right) E_{T-} \\
&\leq \gamma \sum_{i=1}^{m} D_{1+}(i) \exp\left(-y_{i} f(x_{i})\right) \\
&\quad + \left(1-\gamma\right) \sum_{i=m+1}^{n} D_{1-}(i) \exp\left(-y_{i} f(x_{i})\right)\\
&=\gamma\, \tilde{E}_{T+} + \left(1-\gamma\right) \tilde{E}_{T-} = \tilde{E}_{T}
\end{split}
\end{equation}

\begin{figure}
\centering
\includegraphics[width=3.5in]{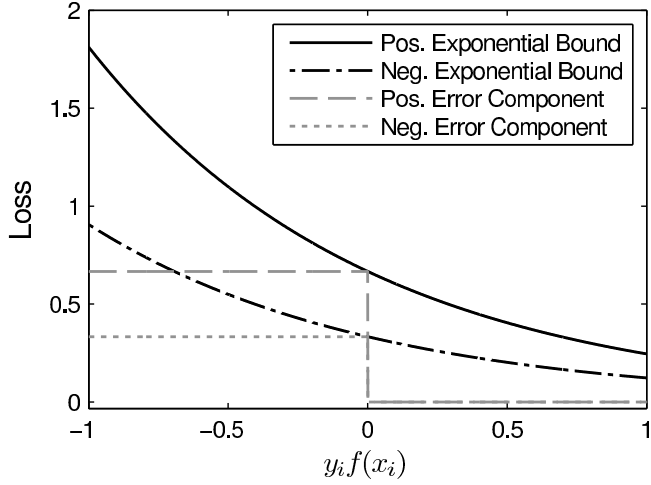}
\caption{AdaBoost training error and its exponential bound split into two class-conditional components for an asymmetry of $\gamma=2/3$.}
\label{classcond_bound_fig}
\end{figure}

In Figure \ref{classcond_bound_fig} we can see the defined weighted partial error bounds ($\tilde{E}_{T+}$ and $\tilde{E}_{T-}$) for an asymmetry of $\gamma=2/3$ (assuming uniform class-conditional distributions). Asymmetry becomes evident.

As it can be seen, the two partial bounds have expressions formally identical to that of the general bound used in the original AdaBoost (\ref{exp_bound2_eqn}), so an equivalent update rule can be derived for each class error:

\begin{gather}
\label{posweight_rule_eqn}
D_{(t+1)+}(i)=
\frac{D_{t+}(i)\exp\left(-\alpha_{t} y_{i} h_{t}(x_{i})\right)}{Z_{t+}} \\
\label{negweight_rule_eqn}
D_{(t+1)-}(i)=
\frac{D_{t-}(i)\exp\left(-\alpha_{t} y_{i} h_{t}(x_{i})\right)}{Z_{t-}}
\end{gather}

where

\begin{gather}
\label{ztpos_eqn}
Z_{t+}=
\sum_{i=1}^{m} D_{t+}(i) \exp\left(-\alpha_{t} y_{i} h_{t}(x_{i})\right)\\
\label{ztneg_eqn}
Z_{t-}=
\sum_{i=m+1}^{n} D_{t-}(i) \exp\left(-\alpha_{t} y_{i} h_{t}(x_{i})\right)
\end{gather}

We will also define two new parameters $P_{t+}$ and $P_{t-}$ which, unraveling the update rules, can be expressed as follows:

\begin{gather}
\label{ptpos_eqn}
P_{t+}=
\prod_{k=1}^{t-1} Z_{k+}=
\sum_{i=1}^{m} \left(D_{1+}(i) \prod_{k=1}^{t-1}\exp\left(-\alpha_{k} y_{i} h_{k}(x_{i})\right)\right)\\
\label{ptneg_eqn}
P_{t-}=
\prod_{k=1}^{t-1} Z_{k-}=
\sum_{i=m+1}^{n} \left(D_{1-}(i) \prod_{k=1}^{t-1}\exp\left(-\alpha_{k} y_{i} h_{k}(x_{i})\right)\right)
\end{gather}

These parameters (we will discuss later about their meaning) allow us to express the partial error bounds in a compact form:

\begin{gather}
\label{pos_et_pt_zt_eqn}
\tilde{E}_{t+}=P_{t+} \, Z_{t+} \\
\label{neg_et_pt_zt_eqn}
\tilde{E}_{t-}=P_{t-} \, Z_{t-}
\end{gather}

The global error bound of the original view of AdaBoost, $\tilde{E}_{t}$, can also be analogously rewritten by defining an equivalent parameter $P_{t}$ for the whole training set:

\begin{gather}
\label{ptpos_gl}
P_{t}=
\prod_{k=1}^{t-1} Z_{k}=
\sum_{i=1}^{n} \left(D_{1}(i) \prod_{k=1}^{t-1}\exp\left(-\alpha_{k} y_{i} h_{k}(x_{i})\right)\right)\\
\label{et_pt_zt_eqn}
\tilde{E}_{t}=P_{t} \, Z_{t}
\end{gather}

As a result, the error bound to minimize can be expressed as:

\begin{equation}
\label{bound_classcond_decomp_eqn}
\begin{split}
\tilde{E}_{t}&=\gamma\, \tilde{E}_{t+} + \left(1-\gamma\right) \tilde{E}_{t-}\\ 
&=\gamma\, P_{t+} Z_{t+} + \left(1-\gamma\right) P_{t-} Z_{t-}
\end{split}
\end{equation}

Bearing in mind that in each round the only variable parameters are $Z_{t+}$ and $Z_{t-}$ ($\gamma$ is fixed from the beginning, and $P_{t}$ depends only on the previous rounds), we can minimize $\tilde{E}_{t}$ using a procedure analogous to that proposed by \citet{SchapireSinger99}. While the minimization is exactly the same as in the original case ($\partial\tilde{E}_{t}/\partial\alpha_{t}=0$) the process can be entirely performed in terms of the class-conditional parameters and allows us to obtain the next expression of the error to be minimized round by round:

\begin{equation}
\label{round_err_classcond_eqn}
\epsilon_{t}=
\frac{\gamma\,P_{t+}} {\gamma\, P_{t+} + \left(1-\gamma\right) P_{t-}} \,\epsilon_{t+}+
\frac{\left(1-\gamma\right) P_{t-}} {\gamma\, P_{t+} + \left(1-\gamma\right) P_{t-}} \,\epsilon_{t-}
\end{equation}

Where $\epsilon_{t+}$ and $\epsilon_{t-}$ are the partial weighted errors per class:

\begin{gather}
\label{pos_round_err_eqn}
\epsilon_{t+}=
\sum_{i=1}^{m} D_{t+}(i) \llbracket h(x_{i}) \neq y_{i}\rrbracket =
\sum_{\textrm{pos nok}} D_{t+}(i)\\
\label{neg_round_err_eqn}
\epsilon_{t-}=
\sum_{i=m+1}^{n} D_{t-}(i) \llbracket h(x_{i}) \neq y_{i}\rrbracket =
\sum_{\textrm{neg nok}} D_{t-}(i)
\end{gather}

The expression for $\alpha_{t}$ is:

\begin{equation}
\label{alphat_classcond_eqn}
\begin{split}
\alpha_{t}&=\frac{1}{2} \\
& \quad \ln \left(\frac 
{\gamma\, P_{t+} \displaystyle \sum_{\textrm{pos ok}} D_{t+}(i) + \left(1-\gamma\right) P_{t-} \sum_{\textrm{neg ok}} D_{t-}(i)}
{\gamma\, P_{t+} \displaystyle \sum_{\textrm{pos nok}} D_{t+}(i) + \left(1-\gamma\right) P_{t-} \sum_{\textrm{neg nok}} D_{t-}(i)} \right) \\
&=\frac{1}{2} \ln \left( \frac{1-\epsilon_{t}}{\epsilon_{t}}\right)
\end{split}
\end{equation}

And the final training error bound, can be expressed as:

\begin{gather}
\label{et_bound_eqn2}
E_{T} \leq \tilde{E}_{T} = \prod_{t=1}^{T} Z_{t} = \prod_{t=1}^{T} \sqrt{1-{r_{t}}^2} \\
\label{rt_classcond_eqn}
\begin{split}
r_{t} &= 
\frac {\gamma\, P_{t+}} {\gamma\, P_{t+} + \left(1-\gamma\right) P_{t-}}
\sum_{i=1}^{m} D_{t+}(i) y_{i} h_{t}(x_{i}) \\
& \quad + \frac {\left(1-\gamma\right) P_{t-}} {\gamma\, P_{t+} + \left(1-\gamma\right) P_{t-}}
\sum_{i=m+1}^{n} D_{t-}(i) y_{i} h_{t}(x_{i})
\end{split}
\end{gather}
\vspace{3pt}

As we can see, all the magnitudes ($\epsilon_{t}$, $\alpha_{t}$ and $r_{t}$) are systematically decoupled in two components according to the global asymmetry and the classifier behavior over each class. The key concept is that expressions (\ref{round_err_classcond_eqn}), (\ref{alphat_classcond_eqn}) and (\ref{rt_classcond_eqn}) are actually the same as those of the original AdaBoost formulation (\ref{round_err_eqn}), (\ref{alphat2_eqn}) and (\ref{rt_eqn}), respectively. On one hand, the derivation is equivalent to the original with the only exception that weights are decomposed in three parameters (\ref{asym_eqn}), (\ref{pos_weight_ini_eqn}), (\ref{neg_weight_ini_eqn}). On the other hand, during the derivation process we can obtain equivalences (\ref{rule1_eqn}), (\ref{rule2_eqn}), (\ref{rule3_eqn}) which appropriately replaced on the original AdaBoost expressions lead us to the new ones.

\begin{gather}
\label{rule1_eqn}
\gamma\, P_{t+}+\left(1-\gamma\right) P_{t-} = P_{t} \\
\label{rule2_eqn}
\gamma\, P_{t+} D_{t+}(i) = P_{t} D_{t}(i), \; \textrm{for } i=1,\ldots,m \\
\label{rule3_eqn}
\left(1-\gamma\right) P_{t-} D_{t-}(i) = P_{t} D_{t}(i), \; \textrm{for } i=m+1,\ldots,n
\end{gather}

\subsection{Asymmetric Error Analysis}
\label{sub_sec:asym_error}

The initial weight decomposition in our analysis allows us to decouple the global weight distribution information in two levels which were always mixed in the original AdaBoost formulation:

\vspace{3pt}
\begin{itemize}
\item \emph{Class level}:  The asymmetry parameter $\gamma$ models the global cost of the positive class over the negative one. From a practical point of view, this parameter can be used to introduce asymmetry in the strong classifier.
\vspace{3pt}
\item \emph{Example level}: The class-conditional initial weight distributions ($D_{1+}$ and $D_{1-}(i)$) model the relative relevance of each example inside its own class. So, being two separate distributions, they are isolated from the asymmetry of the problem.
\end{itemize}
\vspace{3pt}

This two-level categorization can be extrapolated to the error bound minimized by AdaBoost in each round, yielding us a new insight.

\setlength{\arraycolsep}{0.0em}
\begin{equation}
\begin{split}
\label{bound_categorization_eqn}
\tilde{E}_{t}& {}={} \quad \underbrace{\overbrace{\underbrace{\underbrace{\gamma}_{\textrm{Global Asymmetry}}\cdot
\underbrace{P_{t+}}_{\textrm{Previous Rounds Asymmetry}}}_{\textrm{Effective Asymmetry}}}^{\textrm{CLASS LEVEL}} \cdot \overbrace{\underbrace{Z_{t+}}_{\textrm{Current Round}}}^{\textrm{EXAMPLE LEVEL}}}_{\textrm{POSITIVES BEHAVIOR}}\\\\
& \quad {}+{} \underbrace{\overbrace{\underbrace{\underbrace{\left( 1-\gamma \right)}_{\textrm{Global Asymmetry}}\cdot
\underbrace{P_{t-}}_{\textrm{Previous Rounds Asymmetry}}}_{\textrm{Effective Asymmetry}}}^{\textrm{CLASS LEVEL}} \cdot \overbrace{\underbrace{Z_{t-}}_{\textrm{Current Round}}}^{\textrm{EXAMPLE LEVEL}}}_{\textrm{NEGATIVES BEHAVIOR}}
\end{split}
\end{equation}
\setlength{\arraycolsep}{5pt}

The bound consists of two formally identical terms, one per class (positive and negative). Each term has two main components: one on the class level and another one in the example level.

\vspace{3pt}
\begin{itemize}
\item The \emph{class level} defines the effective asymmetry demanded for the current round. It can be seen as the global desired asymmetry modulated by the past asymmetric behavior of the classifier (encoded by cumulative errors $P_{t+}$ and $P_{t-}$). It only depends on the previous rounds.
\vspace{3pt}
\item The \emph{example level} is related to the weighted error of the current weak classifier. Weight distributions ($D_{t+}(i)$ and $D_{t-}(i)$) are updated, round by round, to encode the effective relative relevance of each example totally apart from the class behavior. It depends both on the previous and current rounds.
\end{itemize}
\vspace{3pt}

As we can see, the effective asymmetry of each round will depend on the asymmetry of the previous ones, so AdaBoost goal is to iteratively find the weak hypothesis which, given its predecessors, best helps to the global asymmetry minimizing the training error bound. Asymmetry is reached in a round-by-round adaptive way, without any previous restriction on the final number of rounds.

This error bound interpretation can open the door to new modifications of AdaBoost based, for example, on tuning the global and past asymmetry contributions in order to achieve different asymmetric behaviors along rounds.

\subsection{Algorithm}
\label{sub_sec:algorithm_reform}

Once we have seen the actual asymmetric properties of AdaBoost when using a generic initial distribution, the complete algorithm can be reformatted as in Table \ref{AdaBoost_reform_algorithm_tab}.

\begin{table}
\caption{Discrete AdaBoost generalized formulation for asymmetric classification problems.}
\vspace{15pt}
\label{AdaBoost_reform_algorithm_tab}
\centering
\begin{tabular}{p{5.05in}}
\hline
\\{{\footnotesize Given:}}
\begin{itemize} 
\item {{\footnotesize A set of positive examples: $(x_{i},y_{i})=(x_{1},1),\ldots,(x_{m},1)$}}
\vspace{-8pt}
\item {{\footnotesize A set of negative examples: $(x_{i},y_{i})=(x_{m+1},-1),\ldots,(x_{n},-1)$}}
\vspace{-8pt}
\item {{\footnotesize An asymmetry parameter: $\gamma\in(0,1)$.}}
\vspace{-8pt}
\item {{\footnotesize Two weight distributions over the positive $\left(D_{1+}(i)\right)$ and negative examples $\left(D_{1-}(i)\right)$.}}
\end{itemize}
{{\footnotesize Initialize the global weight distribution as:}}
\begin{itemize} 
\item {{\footnotesize $D_{1}(i)=\gamma \: D_{1+}(i)\quad$for $i=1,\ldots,m$}}
\vspace{-8pt}
\item {{\footnotesize $D_{1}(i)=\left(1-\gamma\right) D_{1-}(i)\quad$for $i=m+1,\ldots,n$}}
\end{itemize}
{{\footnotesize For $t=1,\ldots,T$ (or until the strong classifier reaches some performance goal):}}
\begin{itemize} 
\item {{\footnotesize Select the weak classifier $h_{t}(x)$ with the lowest weighted error
$$\epsilon_{t}=\sum_{i=1}^{n}D_{t}(i) \llbracket h_{t}(x_{i})\neq y_{i} \rrbracket = \sum_{\textrm {nok}} D_{t}(i)$$}}
\vspace{-8pt}
\item {{\footnotesize Calculate
$$ \alpha_{t}=\frac{1}{2}\ln\left(\frac{1-\epsilon_{t}}{\epsilon_{t}}\right)$$}}
\vspace{-8pt}
\item {{\footnotesize Update the weight distribution
$$D_{t+1}(i)= \frac{D_{t}(i) \exp\left(-\alpha_{t} y_{i} h_{t}(x_{i})\right)} {\sum_{i=1}^{n} D_{t}(i) \exp\left( -\alpha_{t} y_{i} h_{t}(x_{i}) \right)}$$}}
\end{itemize}
{{\footnotesize The final strong classifier is:
$$ H(x)=\mathrm{sign}\left(\sum_{t=1}^T\alpha_{t}h_{t}(x)\right) $$}}
\\
\hline
\end{tabular}
\end{table}

The only change regarding to the algorithm description usually found in the literature is that the initial weight distribution is not necessarily uniform. Here, we initialize it in terms of an asymmetry parameter $(\gamma)$ and two class-conditional distributions $\left(D_{1+}(i) \textrm { and } D_{1-}(i)\right)$, which can be uniform (all the examples of each class weight the same) or not (some examples are emphasized).

\subsection{Experiments}
\label{sub_sec:example}

In order to illustrate our analysis with empirical results on the asymmetric behavior of AdaBoost with unbalanced initial weight distributions, we performed three kinds of experiments. For these experiments we have defined the Asymmetric Error (AsErr) as the cost-sensitive error of the classifier: the weighted average of positives (PorErr) and negatives (NegErr) error rates or, what is the same, the weighted average of false negatives (FN) and false positives (FP) rates.

\begin{equation}
\label{asym_error_eqn}
\begin{split}
\mathrm{AsErr}&=\gamma \cdot \mathrm{PosErr} + \left(1-\gamma\right) \cdot \mathrm{NegErr} \\
&=\gamma \cdot \mathrm{FN} + \left(1-\gamma\right) \cdot \mathrm{FP}
\end{split}
\end{equation}

\begin{figure}
\centering
\subfloat[] 
{
    \label{training_set_nonover_fig}
    \includegraphics[width=1.72in]{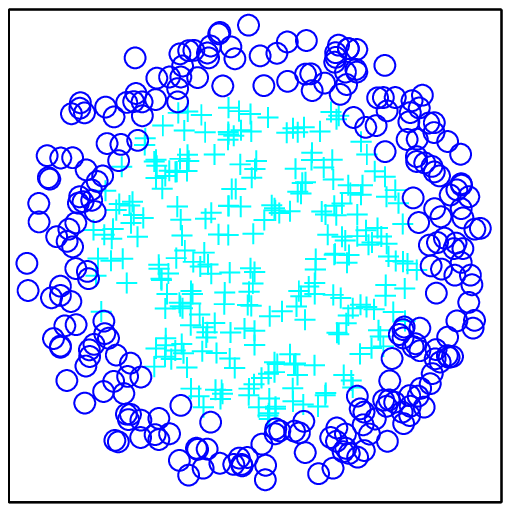}
}
\hspace{1cm}
\subfloat[] 
{
    \label{training_classifiers_nonover_fig}
    \includegraphics[width=1.72in]{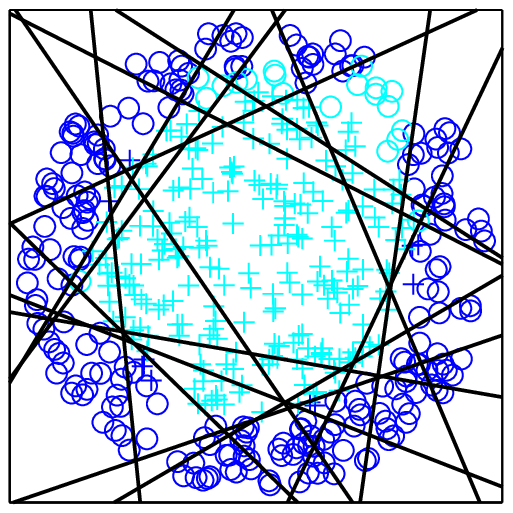}
}
\caption{Training set (a) and example weak classifiers over the test set (b) used to illustrate our asymmetric analysis of AdaBoost. Positive examples are marked as `$+$', while `$\circ$' are the negative ones.}
\label{training_example_nonover_fig} 
\end{figure}

At first, we used the separable set of Figure \ref{training_example_nonover_fig} (inspired by that used by \citealp{ViolaJones02}) in which positives are concentrated in a circular area and negatives surround them, following the same uniform distribution in both cases. Weak classifiers are stumps in the linear two-dimensional space. 

\begin{figure}
\centerline{
{
    \includegraphics[width=1.72in]{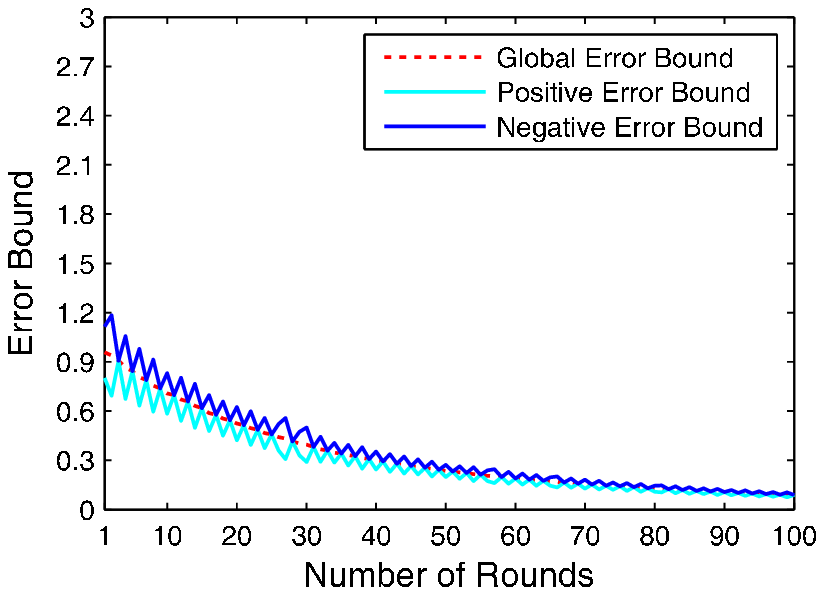}
}
\hfil
\subfloat[$\gamma=\frac{1}{2}$] 
{
    \includegraphics[width=1.72in]{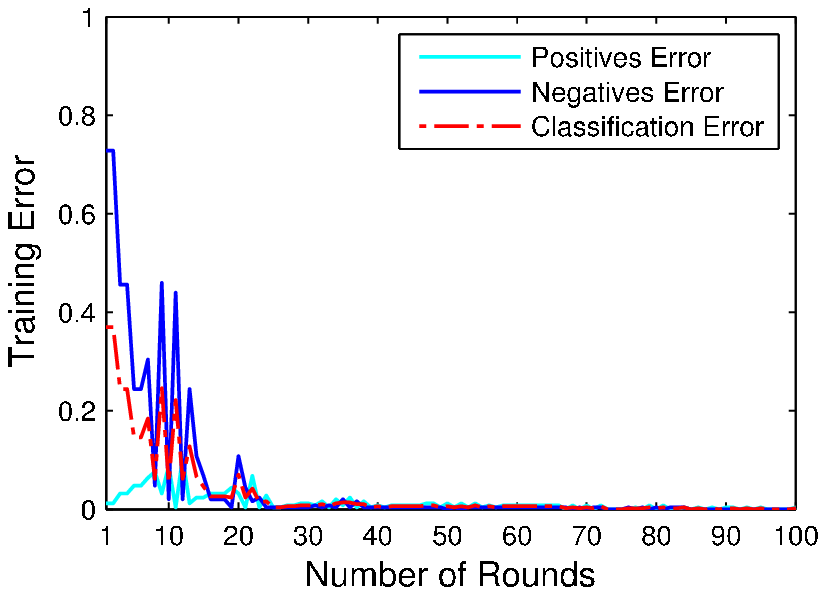}
}
\hfil
{
    \includegraphics[width=1.72in]{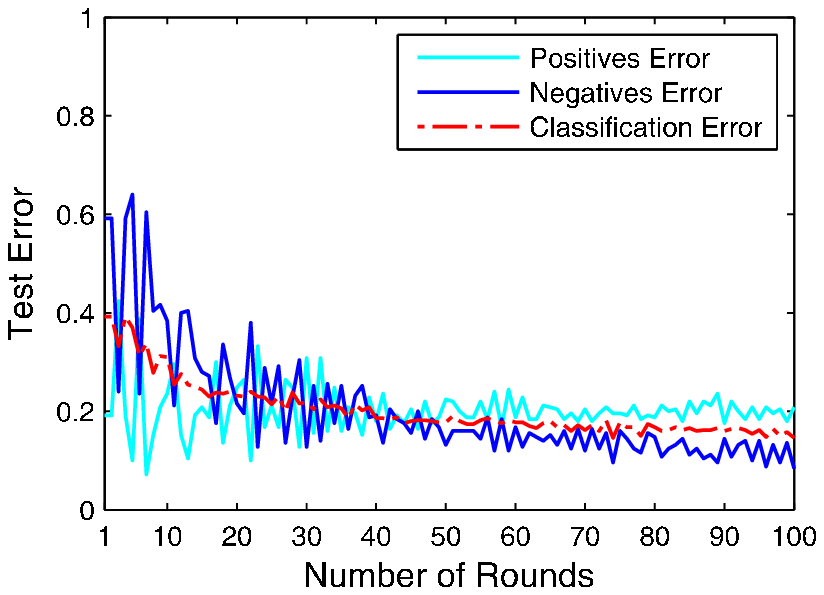}
}
}

\centerline{
{
    \includegraphics[width=1.72in]{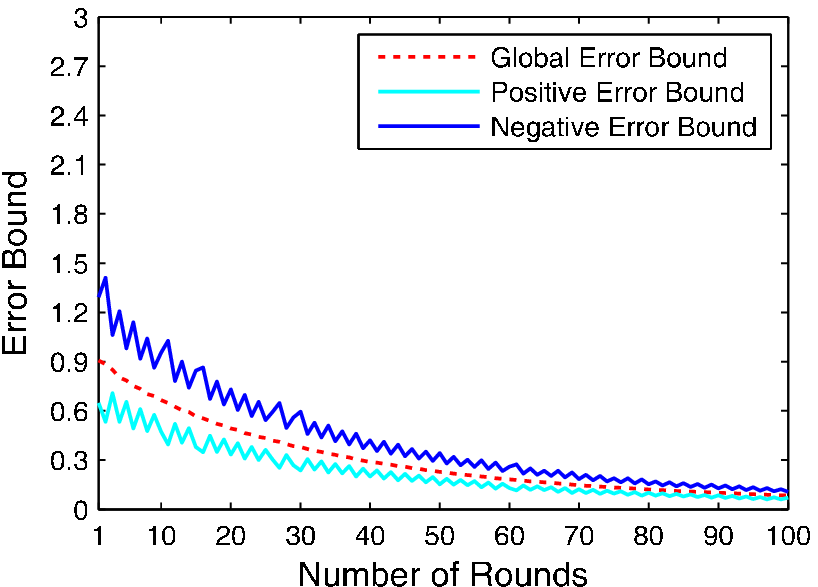}
}
\hfil
\subfloat[$\gamma=\frac{3}{5}$] 
{
    \includegraphics[width=1.72in]{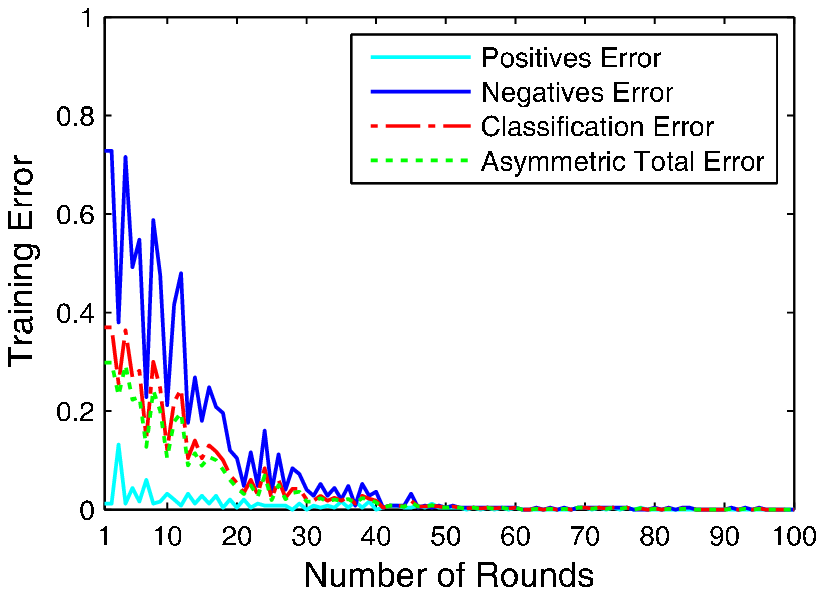}
}
\hfil
{
    \includegraphics[width=1.72in]{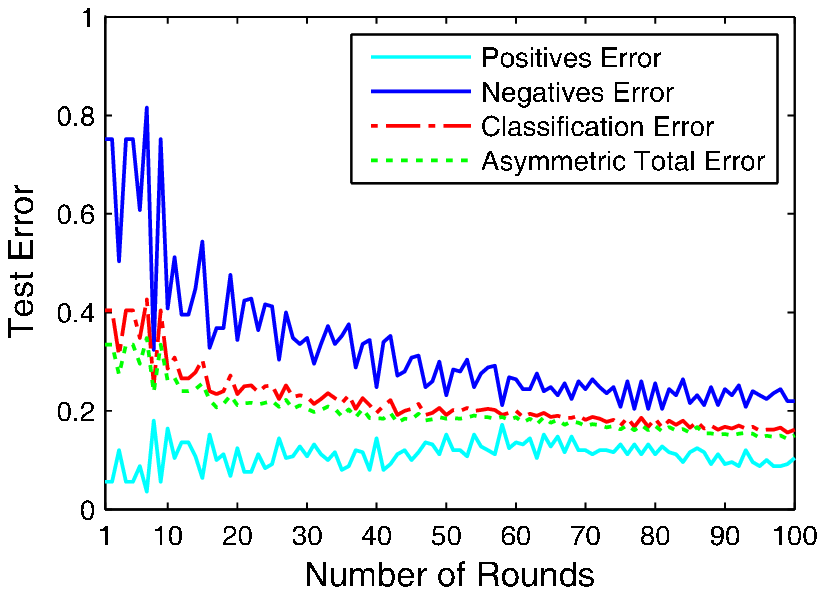}
}
}

\centerline{
{
    \includegraphics[width=1.72in]{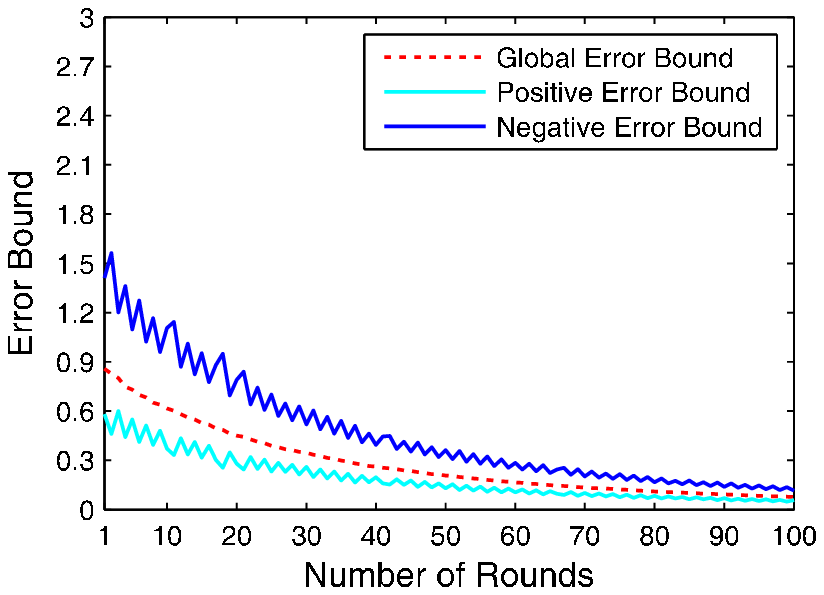}
}
\hfil
\subfloat[$\gamma=\frac{2}{3}$] 
{
    \includegraphics[width=1.72in]{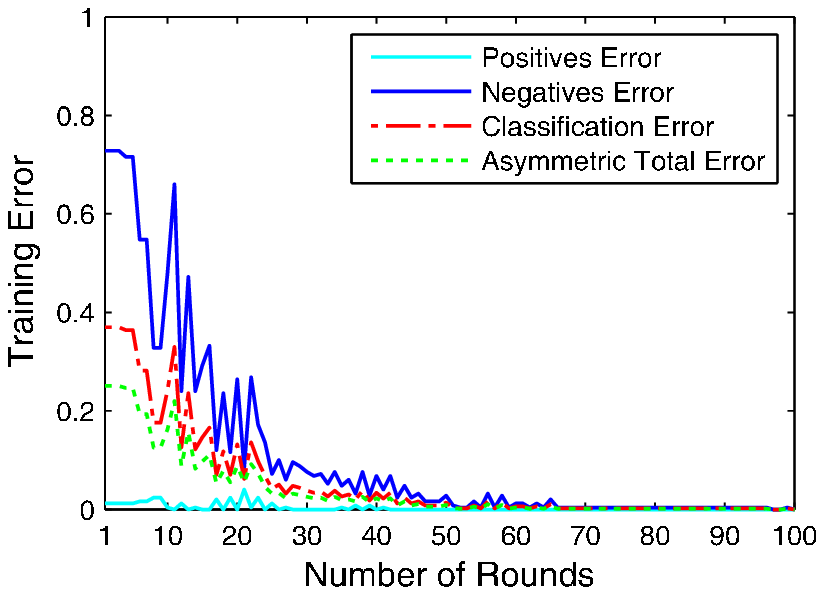}
}
\hfil
{
    \includegraphics[width=1.72in]{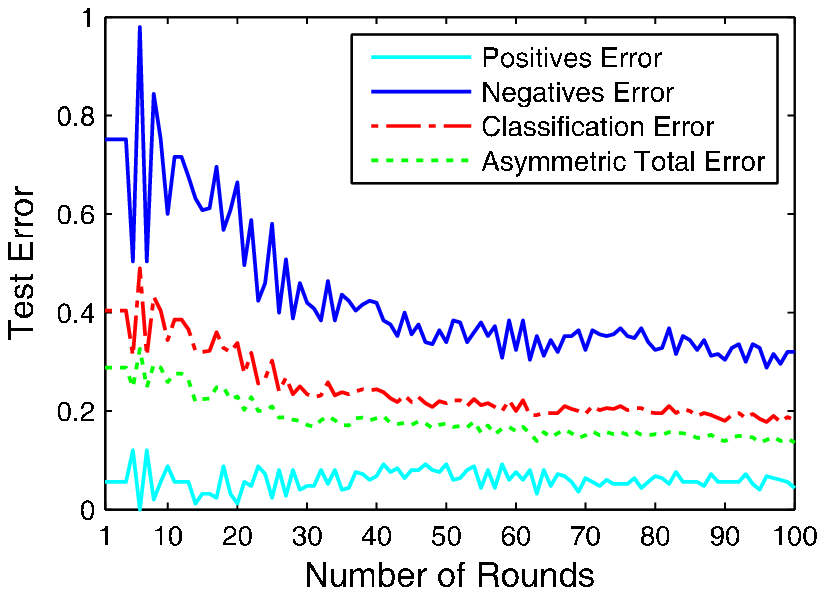}
}
}

\centerline{
{
    \includegraphics[width=1.72in]{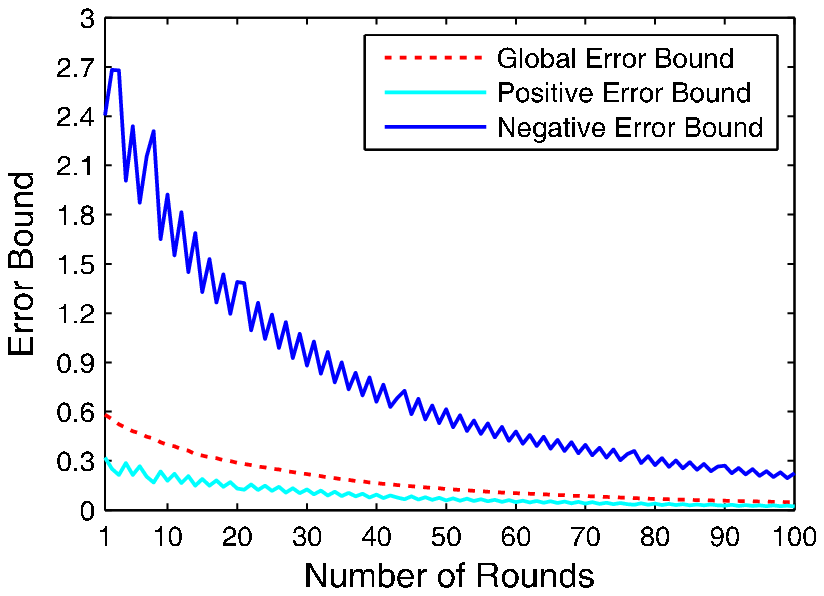}
}
\hfil
\subfloat[$\gamma=\frac{7}{8}$] 
{
    \includegraphics[width=1.72in]{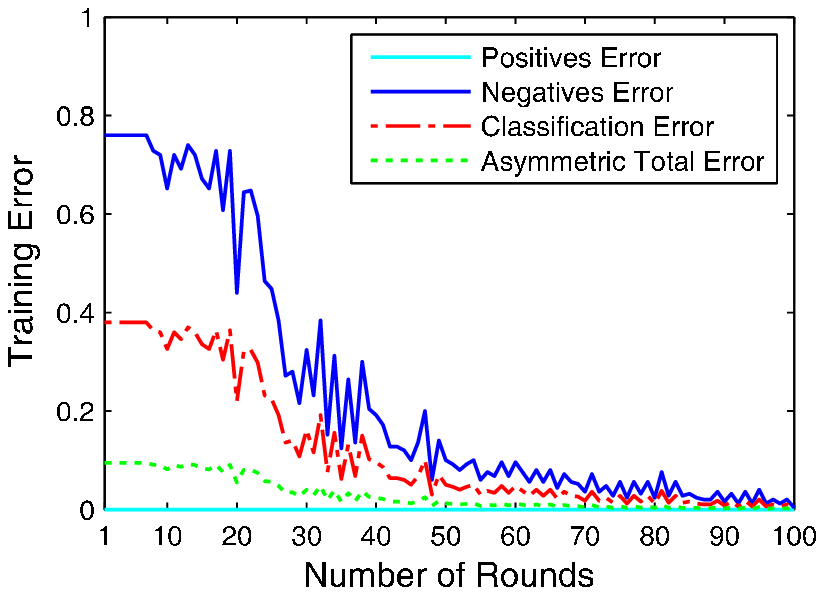}
}
\hfil
{
    \includegraphics[width=1.72in]{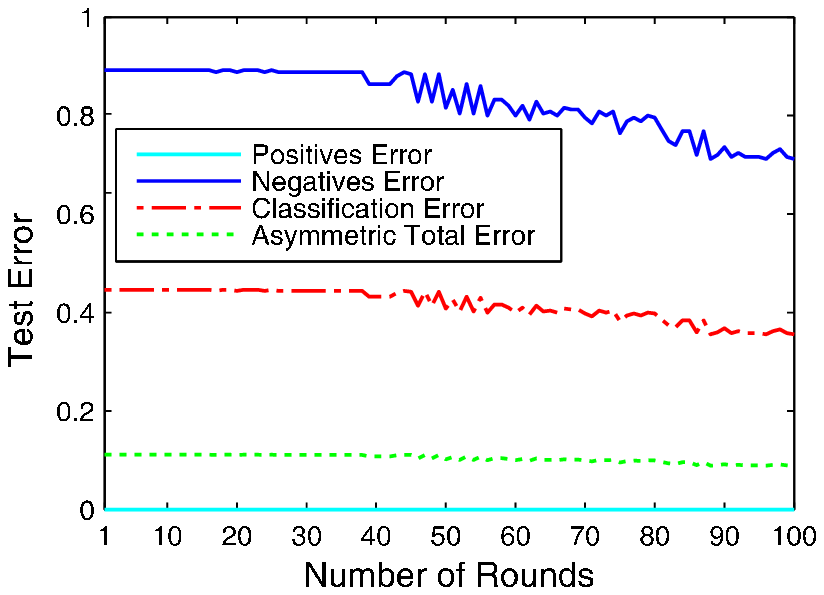}
}
}

\caption{Evolution of training error bounds (left column), training errors (center column) and test errors (right column) through 100 rounds of AdaBoost training and different asymmetries, using the set without overlapping in Figure \ref{training_example_nonover_fig}.}
\label{asym_results_nonover_fig}
\end{figure}

AdaBoost behavior for this training set and different asymmetries ($\gamma=\frac{1}{2}$, $\frac{3}{5}$, $\frac{2}{3}$ and $\frac{7}{8}$) is shown in Figure \ref{asym_results_nonover_fig}. We can see that, as the asymmetry grows, positive error bound and respective positive training/test errors tend to be lower, while negative error bound and respective negative training/test errors tend to be higher. This behavior doesn't prevent the classifier from asymptotically improving itself round by round approaching to zero training error classifiers, due to the separable nature of the classification problem. The key advantage of this approach is that the error evolution follows an unbalanced behavior, allowing the user to stop training at any iteration, with the theoretical confidence of having minimized the error bound with the desired asymmetry no matter in which iteration we are (opposed to Asymboost \citep{ViolaJones02} philosophy). This can be very useful for flexible building of cascaded classifiers as the ones proposed by \citep{ViolaJones04}.

\begin{figure}
\centering
\subfloat[] 
{
    \label{training_set_over_fig}
    \includegraphics[width=1.72in]{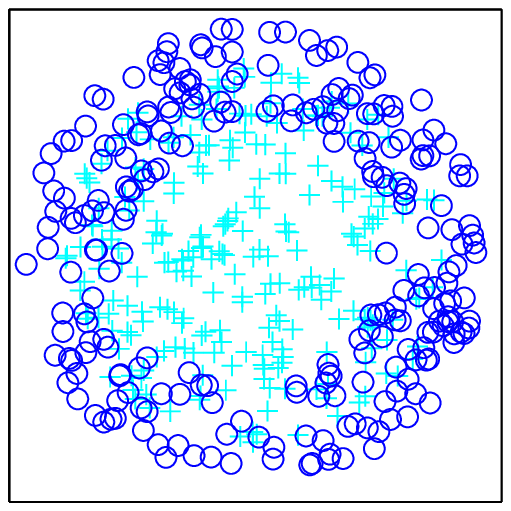}
}
\hspace{1cm}
\subfloat[] 
{
    \label{training_classifiers_over_fig}
    \includegraphics[width=1.72in]{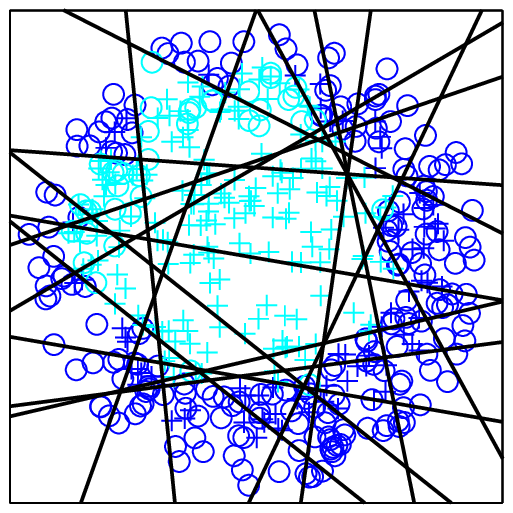}
}
\caption{Training set with overlapping (a) and example weak classifiers over the test set (b). Positive examples are marked as `$+$', while `$\circ$' are the negative ones.}
\label{training_example_over_fig} 
\end{figure}

\begin{figure}
\centerline{
{
    \includegraphics[width=1.72in]{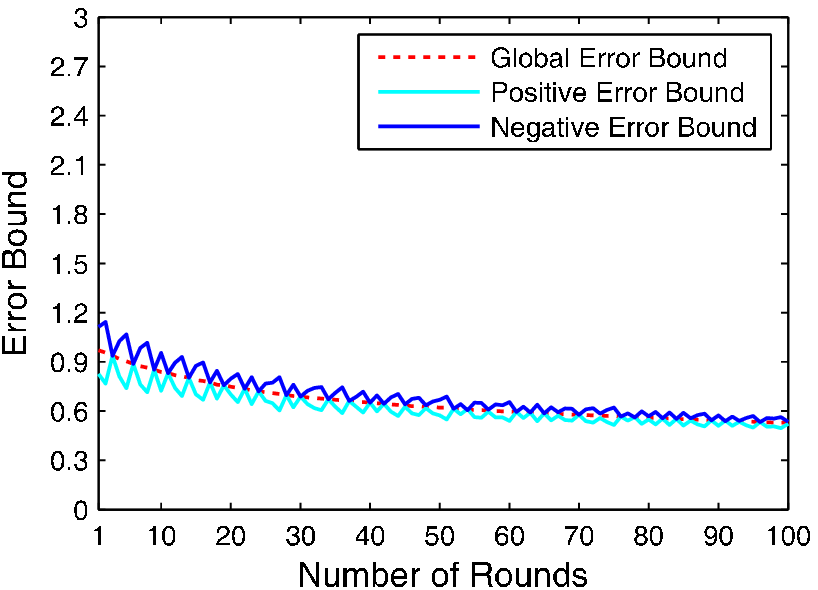}
}
\hfil
\subfloat[$\gamma=\frac{1}{2}$] 
{
    \includegraphics[width=1.72in]{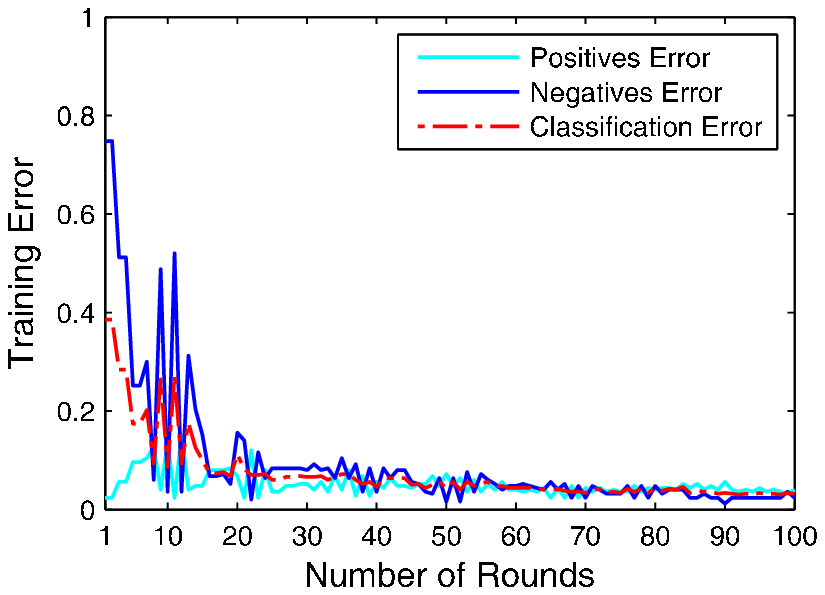}
}
\hfil
{
    \includegraphics[width=1.72in]{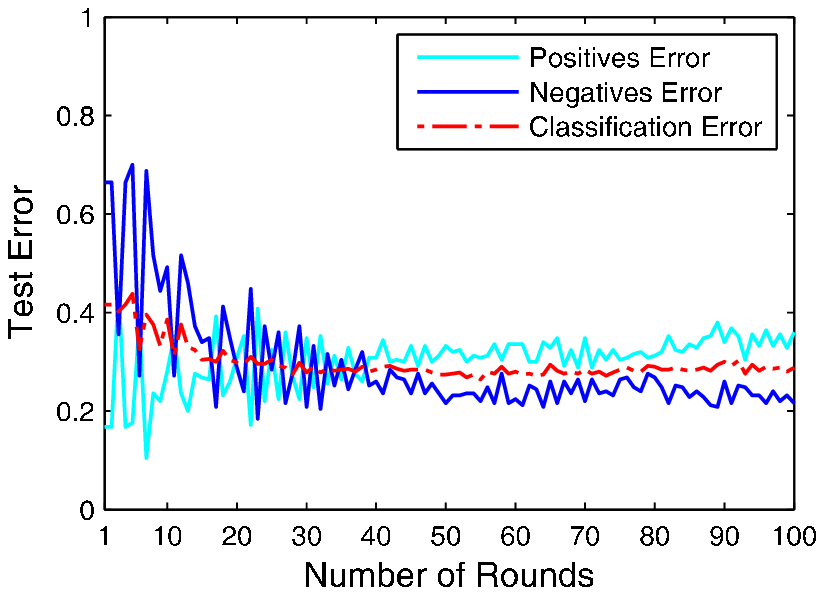}
}
}

\centerline{
{
    \includegraphics[width=1.72in]{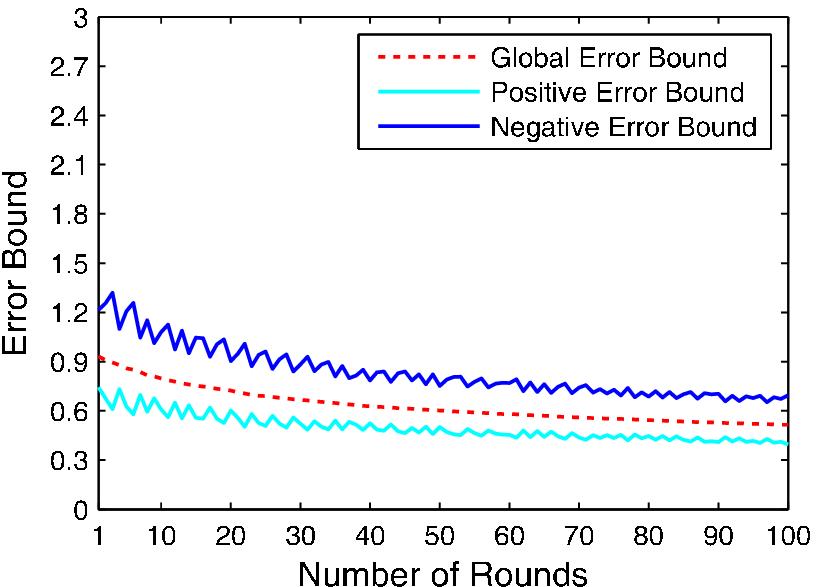}
}
\hfil
\subfloat[$\gamma=\frac{3}{5}$] 
{
    \includegraphics[width=1.72in]{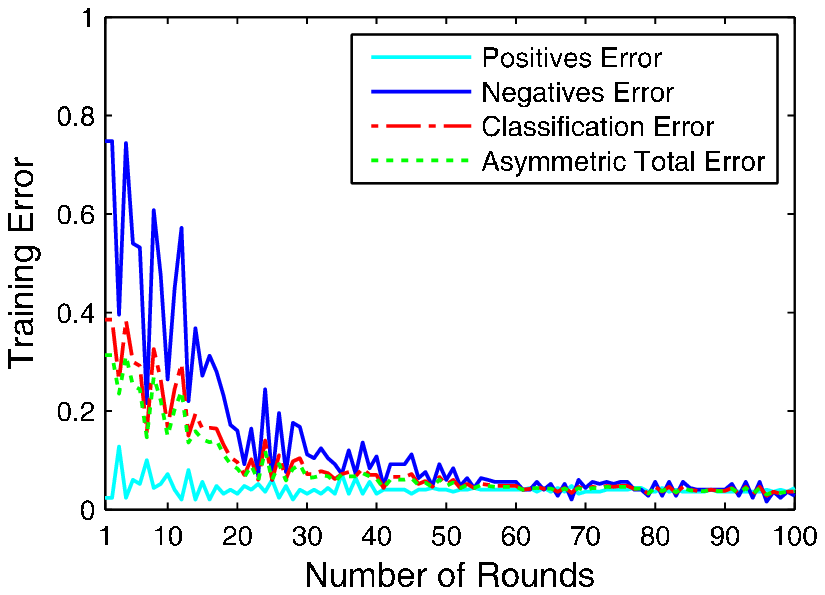}
}
\hfil
{
    \includegraphics[width=1.72in]{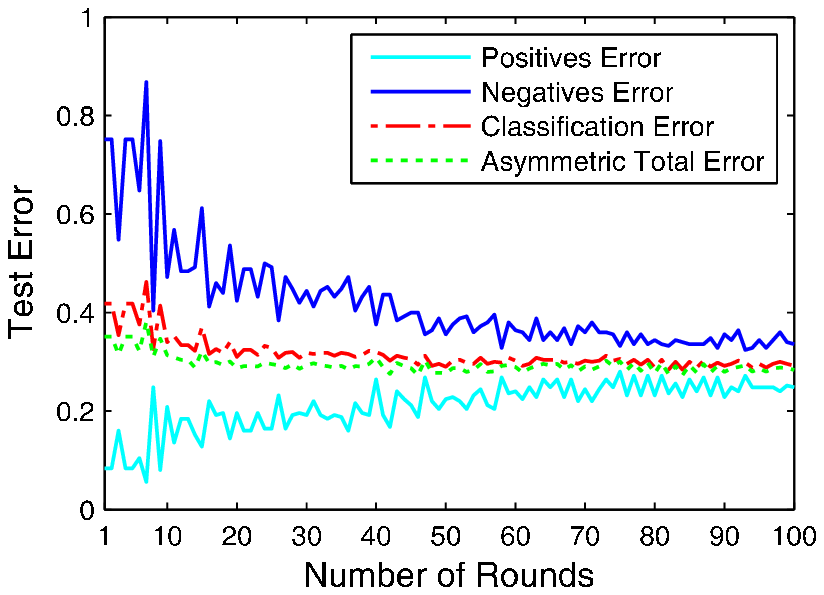}
}
}

\centerline{
{
    \includegraphics[width=1.72in]{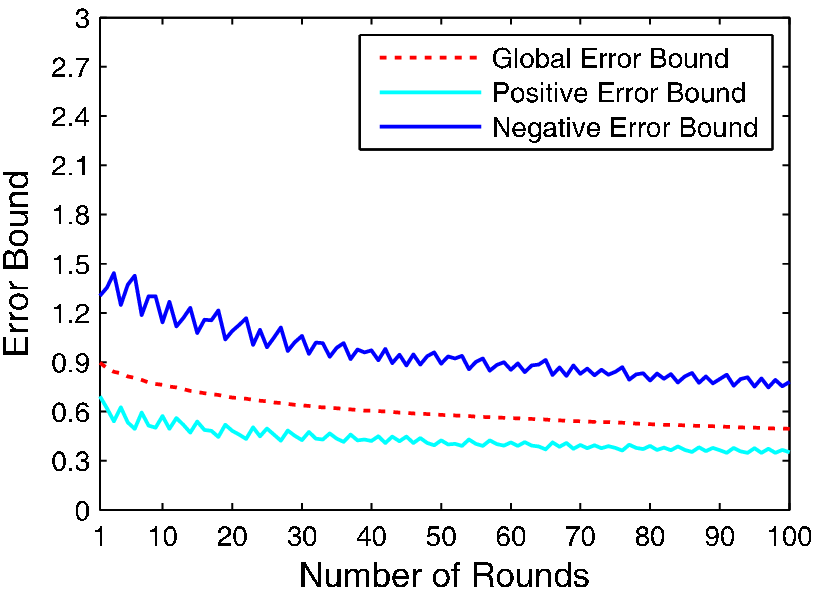}
}
\hfil
\subfloat[$\gamma=\frac{2}{3}$] 
{
    \includegraphics[width=1.72in]{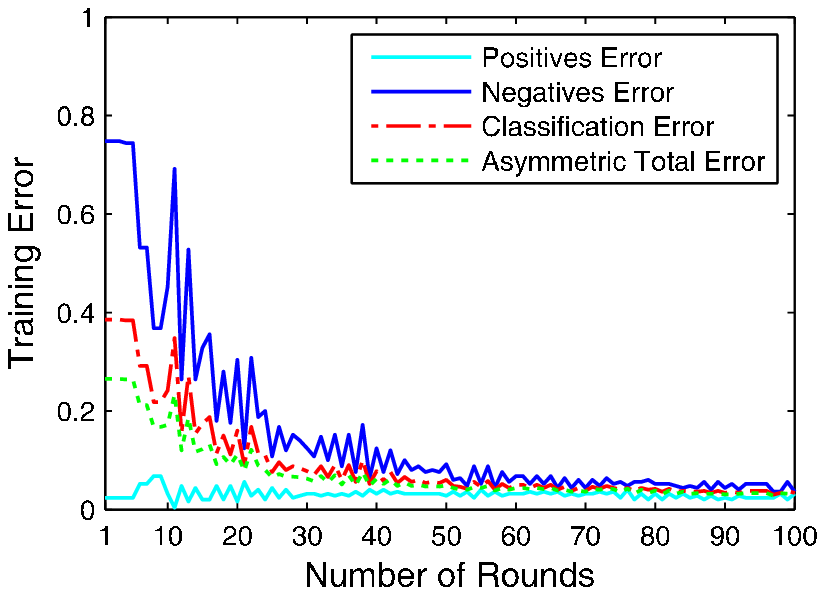}
}
\hfil
{
    \includegraphics[width=1.72in]{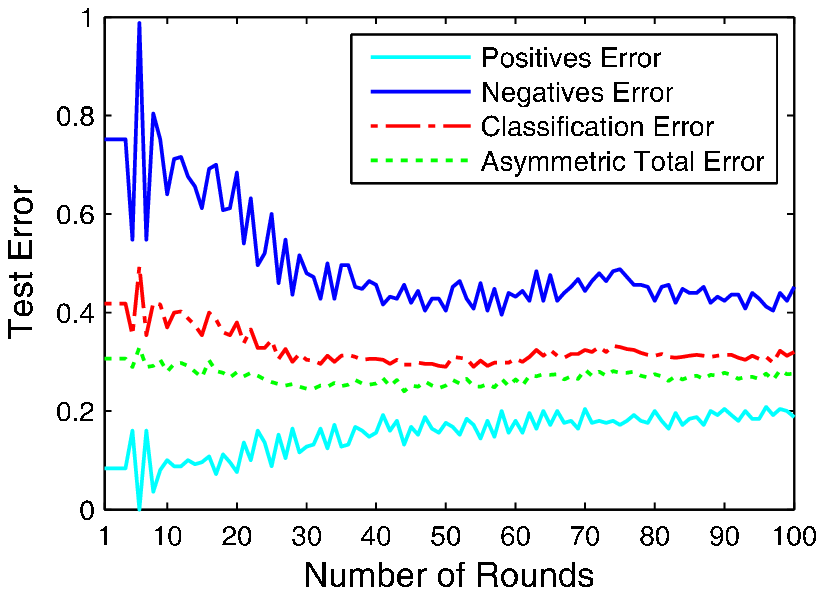}
}
}

\centerline{
{
    \includegraphics[width=1.72in]{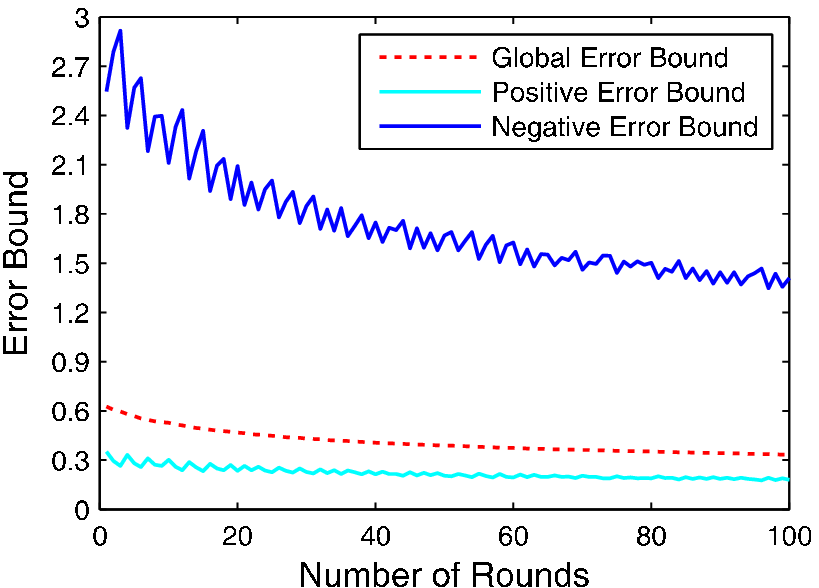}
}
\hfil
\subfloat[$\gamma=\frac{7}{8}$] 
{
    \includegraphics[width=1.72in]{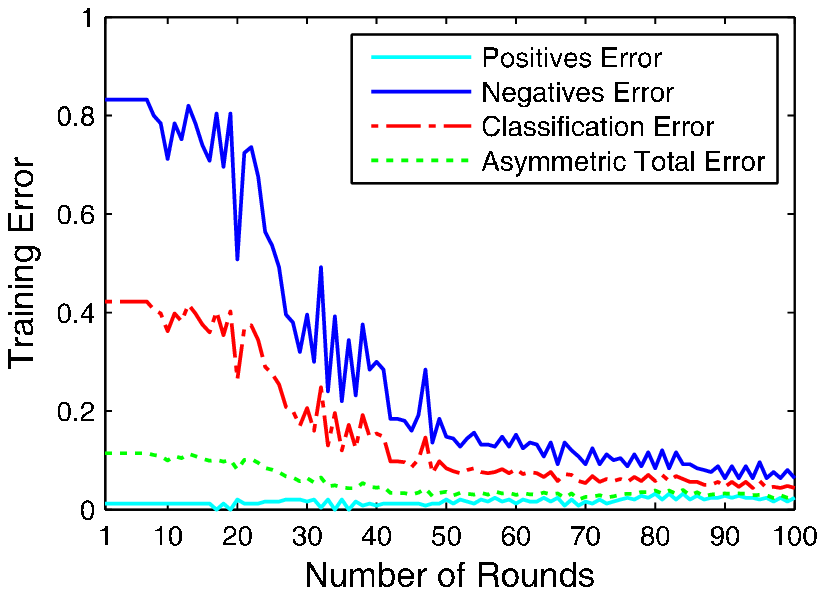}
}
\hfil
{
    \includegraphics[width=1.72in]{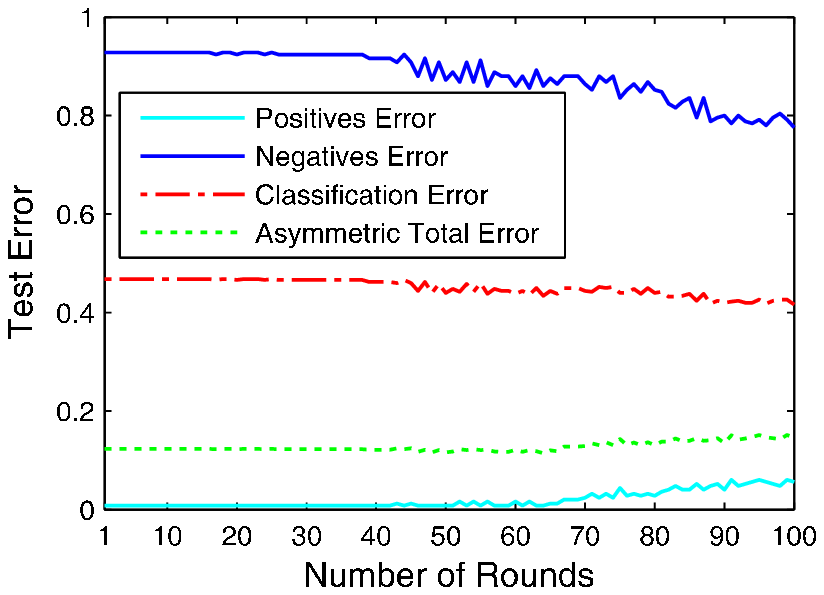}
}
}
\caption{Evolution of training error bounds (left column), training errors (center column) and test errors (right column) through 100 rounds of AdaBoost training and different asymmetries, using the set with overlapping in Figure \ref{training_example_over_fig}.}
\label{asym_results_over_fig}
\end{figure}

\begin{figure}
\centering
\subfloat[$\gamma=\frac{1}{2}$] 
{
    \label{clasresult_05_over_fig}
    \includegraphics[width=1.72in]{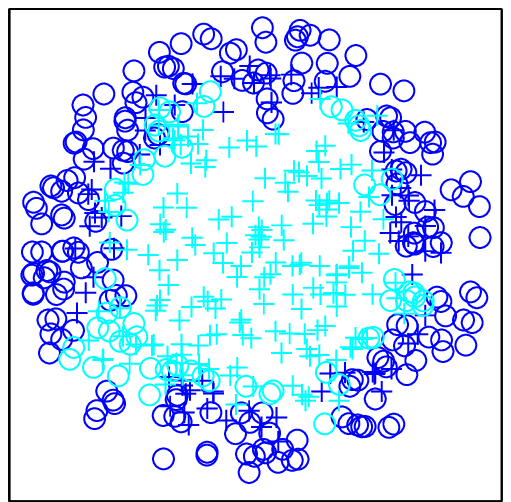}
}
\hspace{0.1cm}
\subfloat[$\gamma=\frac{3}{5}$] 
{
    \label{clasresult_07_over_fig}
    \includegraphics[width=1.72in]{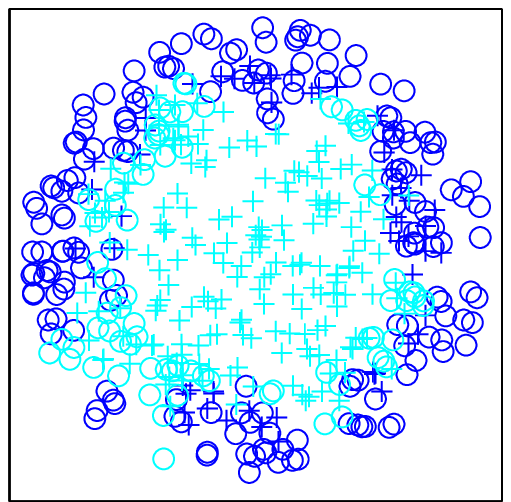}
}
\hspace{0.1cm}
\subfloat[$\gamma=\frac{2}{3}$] 
{
    \includegraphics[width=1.72in]{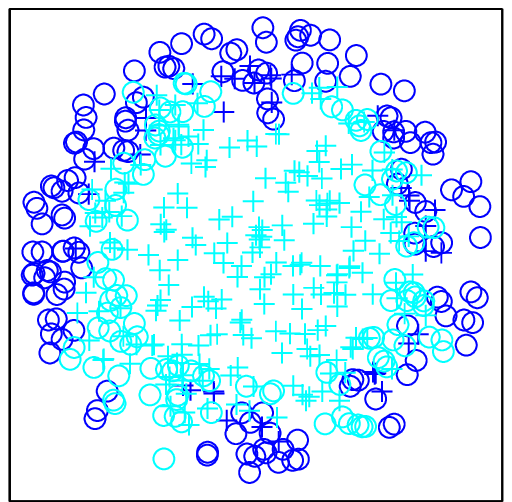}
}
\hspace{0.1cm}
\subfloat[$\gamma=\frac{7}{8}$] 
{
    \includegraphics[width=1.72in]{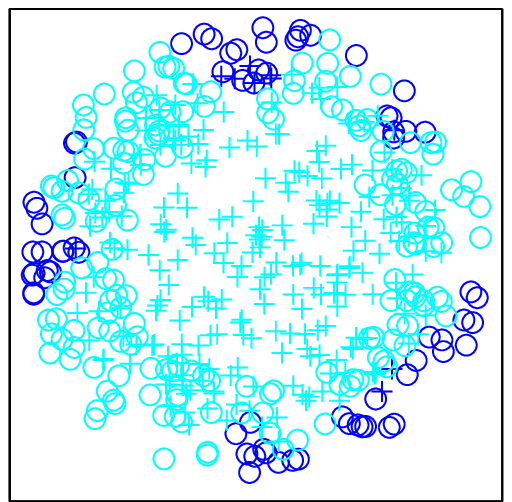}
}
\caption{Classification results over the test set with overlapping (Figure \ref{training_example_over_fig}) for different asymmetries. As in Figure \ref{training_example_nonover_fig} true positives are marked as `$+$', and `$\circ$' are true negatives. However, in this case, cyan colored marks represent positive classifications while blue ones represent negative classifications.}
\label{training_class_over_fig} 
\end{figure}

We also run this experiment with a non-separable set as shown in Figure \ref{training_example_over_fig} and for the same different asymmetries ($\gamma=\frac{1}{2}$, $\frac{3}{5}$, $\frac{2}{3}$ and $\frac{7}{8}$) (Figures \ref{asym_results_over_fig} and \ref{training_class_over_fig}). We can see that, due to the overlapping between classes (they are non-separable), error curves tend to a working point different to that of the previous experiment. In any case, the obtained behaviors are clearly asymmetric along the whole evolution of the boosted classifiers, and the degree of asymmetry is effectively managed by the $\gamma$ parameter.

Finally we have also conducted a more extensive experiment using both synthetic and real datasets to obtain numerical results verifying our hypothesis. The strategy we have followed is \emph{leave-one out cross-validation}. Thus, iteratively selecting every example of a dataset, a classifier is trained over the remaining elements and tested over the selected one. This procedure is repeated for all the examples, all the datasets and all the desired $\gamma$ parameters, so that overall performance figures can be computed. Tables \ref{synth_perform_tab} and \ref{training_example_over_fig} summarize the obtained performance over the synthetic dataset with overlapping in Figure \ref{training_example_over_fig} and some real asymmetric datasets (Credit, Diabetes and Spam) extracted from the UCI Machine Learning Repository \citep{UCIRepository}. As can be seen, in all cases a consistent asymmetric behavior is reached, being also progressive depending on $\gamma$.

\begin{table*}[!htb]
\caption{Classifier behavior (false negatives, false positives, classification error and asymmetric error) for different asymmetric requirements over the synthetic cloud dataset with overlapping in Figure \ref{training_example_over_fig}.}
\scriptsize
\label{synth_perform_tab}
\centering


{\renewcommand{\arraystretch}{1.3} 

\begin{tabular}{|@{}c@{}|@{}c@{}|@{}c@{}|@{}c@{}|@{}c@{}|@{}c@{}|@{}c@{}|}
\hline
\multirow{2}{*}{~$\mathbf{\gamma}$~} & \multicolumn{4}{|c|}{\textbf{Synthetic cloud}}\\
\cline{2-5}
& \textbf{~FN~} & \textbf{~FP~} & \textbf{~ClErr~} & \textbf{~AsErr~~}\\
\hline
{~$\mathbf{1/2}$~} & $~31.60\%~$ & $~29.20\%~$ & $~30.40\%~$ & $~30.40\%~$\\
\hline
{~$\mathbf{3/5}$~} & $26.80\%$ & $38.00\%$ & $32.40\%$ & $31.28\%$\\
\hline
{~$\mathbf{2/3}$~} & $22.00\%$ & $42.00\%$ & $32.00\%$ & $28.67\%$\\
\hline
{~$\mathbf{7/8}$~} & $7.60\%$ & $66.40\%$ & $37.00\%$ & $14.95\%$\\
\hline
\end{tabular}}

\end{table*}

\begin{table*}[!htb]
\caption{Classifier behavior (false negatives, false positives, classification error and asymmetric error) for different asymmetric requirements over real datasets extracted from the UCI Machine Learning Repository \citep{UCIRepository}.}
\scriptsize
\label{uci_perform_tab}
\centering


{\renewcommand{\arraystretch}{1.3} 

\begin{tabular}{|@{}c@{}|@{}c@{}|@{}c@{}|@{}c@{}|@{}c@{}|@{}c@{}|@{}c@{}|@{}c@{}|@{}c@{}|@{}c@{}|@{}c@{}|@{}c@{}|@{}c@{}|}
\hline
\multirow{2}{*}{~$\mathbf{\gamma}$~} & \multicolumn{4}{|c|}{\textbf{Credit}} & \multicolumn{4}{|c|}{\textbf{Diabetes}}  & \multicolumn{4}{|c|}{\textbf{Spam}} \\
\cline{2-13}
& \textbf{~FN~} & \textbf{~FP~} & \textbf{ClErr} & \textbf{AsErr} &\textbf{~FN~} & \textbf{~FP~} &\textbf{ClErr} & \textbf{AsErr} &\textbf{~FN~} & \textbf{~FP~} & \textbf{\,ClErr\,}  &\textbf{\,AsErr\,}\\
\hline
{~$\mathbf{1/2}$~} & $\,28.67\%\,$ & $\,26.86\%\,$ & $\,27.40\%\,$ & $\,27.76\%\,$ & $\,32.09\%\,$ & $\,22.40\%\,$ & $\,25.78\%\,$ & $\,27.24\%\,$ & $\,4.84\%\,$ & $\,6.18\%\,$ & $\,5.37\%\,$ & $\,5.51\%\,$\\
\hline
{~$\mathbf{3/5}$~} & $22.67\%$ & $37.43\%$ & $33.00\%$ & $28.57\%$ & $22.39\%$ & $28.60\%$ & $26.43\%$ & $24.87\%$ & $4.16\%$ & $7.06\%$ & $5.30\%$ & $5.32\%$\\
\hline
{~$\mathbf{2/3}$~} & $18.67\%$ & $43.43\%$ & $36.00\%$ & $26.92\%$ & $19.78\%$ & $32.20\%$ & $27.86\%$ & $23.92\%$ & $3.84\%$ & $8.38\%$ & $5.63\%$ & $5.35\%$\\
\hline
{~$\mathbf{7/8}$~} & $6.00\%$ & $69.14\%$ & $50.20\%$ & $13.89\%$ & $10.07\%$ & $53.00\%$ & $38.02\%$ & $15.44\%$ & $2.33\%$ & $\,11.75\%\,$ & $6.04\%$ & $3.51\%$\\
\hline
\end{tabular}}

\end{table*}

\subsection{Discussion}
\label{sub_sec:discussion}

Previous sections reveal that AdaBoost can be by itself an asymmetric learning algorithm, following its original additive round-by-round updating behavior. Our proposed change of perspective yields several consequences:

\vspace{3pt}
\begin{itemize}
\item The initial weight distribution is more than the distribution seen by the first weak classifier. It is the distribution which weighs the global error bound to be minimized by AdaBoost. Any asymmetry in this initial weight distribution is an effective way to introduce asymmetry in the strong classifier goal.
\vspace{3pt}
\item This kind of asymmetry is asymptotic for the whole classifier and the number of training rounds can be as flexible as in the original case (unlike AsymBoost, \citealp{ViolaJones02}, which rigidly spreads the asymmetry in a predefined number of rounds). Among other advantages, this makes possible, once a strong classifier is trained, to cut it out at whatever round we consider, with the certainty that the error bound has been minimized taking the desired global asymmetry into account. Moreover, it can be specially useful for cascaded classifiers as those used for object detection \citep{ViolaJones04}, in which each stage (each strong classifer) must be markedly asymmetric and as short as possible, in order to improve rejecting efficiency (and consequently the real-time ability of the system).
\vspace{3pt}
\item Asymmetry can be reached without changing the weight update rule, as opposed to the most of the asymmetric AdaBoost modifications in the literature. It is argued that such a modification is needed because AdaBoost updates weights of examples from different classes in the same way, only distinguishing between correctly and incorrectly classified ones. This is true, but it must be taken into account that, before the first weight distribution update, AdaBoost must have selected a first weak classifier $h_{1}(x)$ and a goodness parameter $\alpha_{1}$ according to the initial weight distribution $D_{1}(i)$, which stores the desired asymmetry information. Consequently $h_{1}(x)$ and $\alpha_{1}$ implicitly encode asymmetry information, and both parameters are just the ones that manage the update rule. The result is that asymmetry is indirectly present in the usual weight update rule and, as seen in section~\ref{sub_sec:asym_error}, all the subsequent iterations can be seen as a round-by-round asymmetry adaptive process. Any additional class-dependant change in the weight update rule may emphasize, in a more or less controlled way, the described asymmetric behavior but in those cases it is not clear how it would affect to the theoretical properties of AdaBoost.
\vspace{3pt}
\item The whole formal guarantees provided by AdaBoost remain intact.
\end{itemize}
\vspace{3pt}

\section{Conclusion}
\label{sec:conclusion}

In this paper we have introduced a new insight on the asymmetric learning capabilities of AdaBoost, in which the symmetric case can be seen as a particularization (when asymmetry parameter $\gamma=0.5$). Beyond some preconceptions, the only needed change with regard to the usual formulation is how the initial weights are initialized. We have shown, using a novel class-conditional interpretation of the error bound, that the asymmetric behavior reached is asymptotic with the number of rounds and it works, as the whole algorithm, in an additive round-by-round way. The weight update rule doesn't need to be changed and all the formal guarantees remain intact. Our error bound interpretation can also be useful to develop new AdaBoost modifications based on adjusting the different asymmetry components (both on the class and/or example levels). We have not presented a new algorithm\ldots it is just AdaBoost!


\section{Acknowledgements}
\label{sec:acknowledgements}{This work has been supported by the Spanish Ministry of Science and Innovation through project TEC2008-05894 and by the Galician Government through grants IN840C, IN808C and CN2011/019.}










\vskip 0.2in






\bibliographystyle{model2-names}
\bibliography{Shedding}







\end{document}